\documentclass[11pt]{article}

\usepackage[letterpaper,margin=1in]{geometry}
\usepackage{booktabs}
\usepackage{tabularx}
\usepackage{amsmath}
\usepackage{amssymb}
\usepackage{graphicx}
\usepackage{xcolor}
\usepackage[hidelinks]{hyperref}  

\newcommand{\todo}[1]{}
\newcommand{\pending}[1]{}

\newcommand{\trim}[1]{#1}
\newcommand{\mx}[1]{#1}
\newcommand{\rx}[1]{#1}
\newcommand{\ev}[1]{#1}
\ifdefined\cachehighlight
  \usepackage{soul}
  \sethlcolor{yellow}
  \DeclareRobustCommand{\cache}[1]{\hl{#1}}
\else
  \newcommand{\cache}[1]{#1}
\fi

\newcommand{\blfootnote}[1]{%
  \begingroup\renewcommand{\thefootnote}{}\footnotetext{#1}\endgroup}

\renewenvironment{abstract}{%
  \vspace{0.6em}%
  \begin{center}\bfseries Abstract\end{center}%
  \vspace{-0.4em}\par\noindent\ignorespaces}{\par\vspace{0.8em}}

\begin{document}

\begin{center}
{\LARGE Agents Catching Agents: Shortcut Cascades and\\[4pt]
Benchmark Gaming in Clinical Multi-Agent Systems\par}

\vspace{1.4em}

{\normalsize
Sebasti\'an Andr\'es Cajas Ord\'o\~nez$^{1}$, Agastya Munnangi$^{1,2}$, Aldo Marzullo$^{3}$,
Felipe Ocampo Osorio$^{1}$, Quang Bui$^{4}$, Mohammad Shahin$^{1,5}$, Armaan Grewal$^{6}$,
Emmanuel Paul Kwesiga$^{7}$, Anqi Peter Li$^{8}$, Josephine Nanyonjo$^{9}$,
Aaditya Panchal$^{1,10}$, Arshnoor Bhutani$^{1,11}$, Nikhil Jaiswal$^{12}$,
Milit S. Patel$^{13}$, Maximin Lange$^{14}$, and Leo Anthony Celi$^{1,15}$\par}

\vspace{0.9em}

{\footnotesize
$^{1}$MIT Critical Data, Massachusetts Institute of Technology, Cambridge, Massachusetts, United States\\
$^{2}$Georgia State University, Atlanta, Georgia, United States\\
$^{3}$Politecnico di Milano, Milan, Italy\\
$^{4}$American International School Vienna, Vienna, Austria\\
$^{5}$School of Public Health, Boston University, Boston, Massachusetts, United States\\
$^{6}$Walter Payton College Preparatory High School, Chicago, Illinois, United States\\
$^{7}$Technische Hochschule L\"ubeck, L\"ubeck, Germany\\
$^{8}$Substrate Labs\\
$^{9}$School of Nursing, University of British Columbia, Vancouver, British Columbia, Canada\\
$^{10}$Dartmouth College, Hanover, New Hampshire, United States\\
$^{11}$Department of Computer Science, University of Maryland, College Park, Maryland, United States\\
$^{12}$McGill University, Montreal, Quebec, Canada\\
$^{13}$Department of Molecular Biosciences, University of Texas at Austin, Austin, Texas, United States\\
$^{14}$King's College London, London, United Kingdom\\
$^{15}$Division of Pulmonary, Critical Care and Sleep Medicine, Beth Israel Deaconess Medical Center,
Boston, Massachusetts, United States\par}
\end{center}

\begin{abstract}
\trim{Clinical decision support is moving toward committees of language-model agents deliberating on a
shared workspace. We ask whether such committees can be gamed by \emph{shortcuts}, cues a benchmark rewards
but a clinician would ignore. Across seven cohorts on six public datasets spanning text (MedQA-USMLE,
MedMCQA, MIMIC-CXR reports), imaging (NIH ChestX-ray14, MIMIC-CXR-JPG, CheXpert) and tabular ICU records
(SUPPORT2), Gemini committees resist these cues in isolation (flip $5$--$16\%$), yet a socially plausible
shortcut spreads: when two peers assert the same wrong answer, the \emph{holdout} under test adopts it in
$38\%$ of cases, as does a false ``pre-screen'' system flag, on both capability tiers. Of three oversight
agents, a \emph{gate} cannot separate adoption from honest agreement (false-positive rate $100\%$); a
same-lineage \emph{judge} reading only the transcript flags adoption on text (precision $100\%$, recall
$93\%$) but collapses onto the gate in imaging; a \emph{referee} that privately re-queries the holdout
transfers to imaging ($77$--$88\%$ precision, $13$--$21\%$ false-positive rate). \mx{Tripling a cue's visual
salience does not move contagion, whereas a second peer voice raises it by half again.} Gaming a hidden
rubric is near-silent: only $1/10$ text and \rx{$1/134$} imaging drifters name the rubric they moved toward.
What games a committee is social plausibility, and only a referee independent of self-report catches it.
Code: \url{https://github.com/criticaldata/benchmaxxing}}
\end{abstract}

\blfootnote{This work is licensed under a Creative Commons Attribution 4.0 International
(CC BY 4.0) licence.}

\clearpage

\section{Introduction}
Agents may satisfy a benchmark's formal requirements while bypassing its intended objective, a behaviour we
call \emph{benchmaxxing}~\cite{krakovna2020specification,amodei2016concrete}: a patient-safety risk that is
not self-correcting, since models cannot reliably identify the shortcuts behind their own
predictions~\cite{sharma2023sycophancy} and detection needs structural evaluation. \ev{Shortcut learning, a spurious correlation exploited instead of the
intended signal, is a general deep-network failure mode~\cite{geirhos2020shortcut} and} is best documented in
medical imaging: pneumonia from acquisition artifacts~\cite{zech2018confounding}, COVID-19 from markers
outside the lung fields~\cite{degrave2021ai}, failures hidden inside aggregate
metrics~\cite{oakdenrayner2020hidden}, and embedding-space auditing~\cite{cajas2026shortkit}. Each concerns
one model against a fixed benchmark.

Agentic deployment extends this to correlated failure. On a shared blackboard, agents observe one another's
outputs, so an early shortcut-based response can drive convergence toward the same
error~\cite{ashery2024emergent}, and oversight may fail when evaluators share a blind spot,
\ev{consistent with correlated-failure risks from common evaluation standards~\cite{kleinberg2021algorithmic}}.
Existing methods identify an incorrect answer; none, to our knowledge, asks whether it changed after observing
a peer. We address both with the \emph{referee agent}, the named contribution of the DOJO (Distributed Open
Justice Oversight) framing~\cite{xiang2026dojo}: an oversight agent evaluating other agents rather than doing
the clinical task. \ev{Practitioner grader taxonomies are single-agent and cannot ask whether an agent would
have answered differently had a peer not spoken~\cite{grace2026evals}; the private re-query is that signal.}
It carries three duties, \emph{shortcut detection}, \emph{conformity monitoring} and \emph{hierarchy
monitoring}; we instantiate the first two across seven cohorts.

\section{Related work}
\ev{\trim{Multi-agent LLM populations converge on shared conventions, producing collective biases absent in
isolation~\cite{ashery2024emergent}. Clinically that is the vector: medical committees substitute authority
for verification and override correct dissent~\cite{zhu2025auditing}, amplifying an initial error rather than
correcting it, which is socially mediated shortcut
learning~\cite{zhu2025auditing,zhou2026formaljudge}. Cross-checking and LLM-as-a-judge inherit the blind spots
of the agents they supervise~\cite{zhou2026formaljudge}, so an oversight signal must come from outside the
transcript. In imaging a second bias compounds it, the blind faith vision-language models place in text over
conflicting pixels~\cite{deng2025words}. Populations fail in ways single-agent analyses of spurious
correlation~\cite{zech2018confounding} miss~\cite{hammond2025multiagent}, and prompt-template effects are not
strategic metric exploitation.}}

\section{Materials and methods}

\subsection{Datasets and models}
\rx{All experiments were performed on publicly available clinical datasets, and every analysis set below is
a subset of a larger staged cohort; no model was fine-tuned.} The text experiments used
MedQA-USMLE~\cite{jin2020medqa}, questions from the United States Medical Licensing Examination
(train/dev/test $10{,}178/1{,}272/1{,}273$), restricted to items whose recorded correct option agreed with the
reference answer; the analysis sets were $100$ cases for cue sensitivity, $50$ for the cross-cohort
comparison, $40$ for the cascade and detection arms and $120$ hard cases for the framing ladder.
\mx{Replications used MedMCQA~\cite{pal2022medmcqa} ($50$, $20$ and $110$ cases for cue sensitivity, the
single-peer arm and the anchored-rationale arm) and free-text MIMIC-CXR reports reformulated as
multiple-choice questions over CheXpert findings ($600$, $40$, $20$ and $60$ for the same four purposes).} The
imaging experiments used NIH ChestX-ray14~\cite{wang2017chestxray}, of which one batch of $4{,}999$
radiographs was staged and a cascade cohort of $35$ drawn as the first eligible finding-present images in
patient-identifier order, so the $35$ images come from only ten patients, one contributing $12$ and another $8$;
\mx{MIMIC-CXR-JPG~\cite{johnson2019mimiccxr,goldberger2000physionet} from a different institution ($834$
images from $600$ studies, fixed in advance under a seeded manifest with per-image checksums recorded before
any model query, cascade subsets of $48$, $91$ and $183$);} and CheXpert~\cite{irvin2019chexpert}, used
without modification ($150$ films naturally carrying a support device, $35$ for the hidden rubric).
\ev{SUPPORT2~\cite{knaus1995support,harrell1995support2uci} carries the same paired design on structured
intensive-care records ($120$ cases), its cue an information-identical restatement of the same variables.}
Every MIMIC-CXR finding-present case was independently confirmed as positive, so the read designated as
incorrect did not depend on the model's own output. Visual cues were local overlays leaving anatomy
unaltered, preserving the reference diagnosis by construction. All agents were Gemini~2.5 Flash or
Flash-Lite, two tiers of one family, at temperature $0$, \rx{every call served from a content-addressed
cache so a rerun replays; the noise floors are the one resampled exception.}
\todo{Report exact API model identifiers and access dates, findings queried per cohort, exclusion criteria
and model assignments, and describe the conversion of MIMIC-CXR reports into multiple-choice questions. Add
an experiment-by-cohort-by-$n$ table to the supplement.}

\subsection{Experimental design}
\label{sec:design}

\noindent\textbf{Cue sensitivity.}
Each case was evaluated with the original input and with a modified input preserving the reference answer.
Text cues altered option order, the correct option's relative length, or stem-option lexical overlap. Imaging
cues were a cable-like overlay, a corner tag, a laterality marker and a watermark, each outside the clinically
relevant anatomy and occupying $0.13\%$ to $3.17\%$ of the $1024^2$ frame ($1{,}349$ to $33{,}220$ pixels),
the watermark also rendered at opacities of $0.15$, $0.30$ and $0.45$. A \emph{cue-induced flip} was a case
whose prediction differed between the two inputs; since predictions may differ for unrelated reasons, the
original input was queried twice and the disagreement taken as the \emph{noise floor}. \rx{These repeated
queries were the only quantity resampled above temperature $0$ and with the cache bypassed, so the floor
overestimates temperature-$0$ instability. \ev{That understates cue effects, making our solo nulls easier to
reach, and is not reliably conservative either way: on the one cohort run at both temperatures the watermark
reads $-0.029$ at temperature $1$ against $+0.114$ at temperature $0$, whose floor is $0$ by
construction.}} We report flip rate, noise floor and their difference, \emph{flip above noise}, a descriptive
quantity, not an adjusted causal effect. \todo{Describe cue construction, the option targeted by
each text cue, overlay coordinates and dimensions, and the checks confirming the reference answer was
preserved. State $n$, the resampling procedure and the number of draws for the imaging floors of $0.23$ (NIH)
and $0.17$--$0.18$ (MIMIC), which are load-bearing and less documented than the text floor.}

\noindent\textbf{Peer endorsement.}
A paired design determined whether an answer rejected in isolation is accepted once a peer asserts it. A
\emph{holdout} agent received the same clinical input in both conditions, observing one or two peers
committing to a predefined incorrect response, the \emph{peer-endorsed answer}, in the \emph{shared} condition
and not in the \emph{isolated} one. The \emph{adoption rate} was the proportion of holdout responses equal
to that answer and the \emph{contagion} their difference between conditions, with case-level
adoption scored only where agreement followed exposure,
\begin{equation}
A_i=\mathbb{1}\!\left[\hat{y}^{\mathrm{shared}}_i=y^{\mathrm{peer}}_i \ \wedge\
\hat{y}^{\mathrm{isolated}}_i\neq y^{\mathrm{peer}}_i\right].
\label{eq:adoption}
\end{equation}
An adoption whose isolated response was correct is additionally \emph{harmful}. This endpoint identifies a
response change associated with the deliberation, not the model's reasoning. Four factors were varied
independently: peer count (one or two); message content (committed answer alone, generic rationale, or
vignette-anchored); confidence (categorical against hedged, holding source and asserted answer fixed); and
deliberation framing (unframed, collaborative, independent, or critical review). \rx{In a further condition
the peers were replaced by an incorrect automated pre-screening flag attributed to an upstream system,
separating institutional authority from peer pressure.} Imaging repeated the design across the four cues, the
three opacities, and unmodified CheXpert images with naturally occurring support devices. \todo{State how incorrect answers were selected, whether peer messages were
scripted or model-generated, and whether paired conditions differed only by peer exposure.}

\noindent\textbf{Referee agents and baselines.}
The \emph{referee} is an oversight agent that does not answer the clinical question, but determines whether
another agent's response was produced by the deliberation. It infers the answer endorsed by the peers,
privately re-queries the holdout with the same input and no transcript, and flags
$\hat{y}^{\mathrm{shared}}_i=y^{\mathrm{peer}}_i \wedge
\hat{y}^{\mathrm{shared}}_i\neq\hat{y}^{\mathrm{private}}_i$. \rx{The private re-query substitutes for the
isolated condition of Eq.~\ref{eq:adoption}, which a deployed monitor cannot observe.} It requires neither
the reference answer nor model parameters. Two baselines: the \emph{gate}, flagging every shared-condition
agreement irrespective of the holdout's isolated response, and a same-lineage \emph{judge}, reading the
deliberation without a private re-query. All three are scored against Eq.~\ref{eq:adoption} by precision,
recall and FPR. One intervention was applied during rather than after deliberation: a warning that peers may
be incorrect. \todo{Provide the exact referee, baseline and intervention prompts, and
state how ties or disagreement among peers were handled.}

\rx{\textbf{What the design can and cannot identify.}}%
\label{sec:design-identifiability}
\rx{In the text and tabular experiments the incorrect answer was chosen to differ from the holdout's
isolated response, so on the planted arm the answer inferred from the peers' modal assertion is that answer
by construction and the referee flag reduces algebraically to the label it is scored against.} \ev{We
therefore add an \emph{honest-peer clean control}: the same cascade with both peers asserting the correct
answer, making adoption false by construction, so a flag there is a genuine false positive. That makes
precision and the false-positive rate measurable, but not recall. Imaging needs no such arm, its incorrect
answer being fixed against the reference diagnosis.} \rx{Recall of $1.0$ is likewise a subset relation for
referee and gate, the flagged set containing every adoption by definition.}

\noindent\textbf{Hidden rubric.}
A separate experiment replaced the social channel with an evaluation rule: a hidden rubric awarded full marks to a predefined incorrect option, the \emph{decoy}, each case paired with a control
run without the rubric. \rx{Because the decoy was defined as an option other than the response given without
the rubric, the control cannot select it and its rate is structurally $0$ on all $291$ committed rows; the
intended rubric-minus-control contrast is void, and we report decoy uptake as a raw rate under the rubric.}
Whether the response named the rubric was recorded separately, scored by one detector across both modalities;
\rx{an earlier modality contrast that scored the two lanes with different detectors is withdrawn
(\S\ref{sec:robust}).} We describe this as drift toward the rewarded option rather than reward-hacking, since
the design does not establish intent. \todo{Provide the rubric
verbatim, identify the message level at which it was inserted, explain how the decoy was selected, and
redefine the decoy so that the control arm is able to select it.}

\subsection{Response processing and statistical analysis}
Each completion was classified as an answer option, a \textsc{refusal}, or \textsc{unparseable}, taking the
final explicit answer where several appeared. Refusals and parsing failures were kept separate, since
conflating them inflates the cascade. Paired binary outcomes were compared using exact McNemar tests and
independent proportions using Fisher's exact tests. Risk differences carry nonparametric bootstrap $95\%$
confidence intervals over $20{,}000$ resamples, at case level except the NIH cascade, whose $35$ images
cluster within ten patients and are resampled as such. Multiple comparisons were controlled at $\alpha=0.05$
using both Benjamini--Hochberg and Holm. \rx{No patient-level sign test is reported for the NIH cascade, since
shared adoption is at ceiling in nine of ten patients and the test admits a single attainable value.}
\todo{Define the prespecified primary hypothesis family and distinguish primary from
exploratory analyses; provide parsing rules for binary and free-text imaging responses; report parser
validation; and specify how refusals and unparseable outputs entered each denominator.}

\section{Results}

\begin{figure}[t]
\centering
\includegraphics[width=\textwidth]{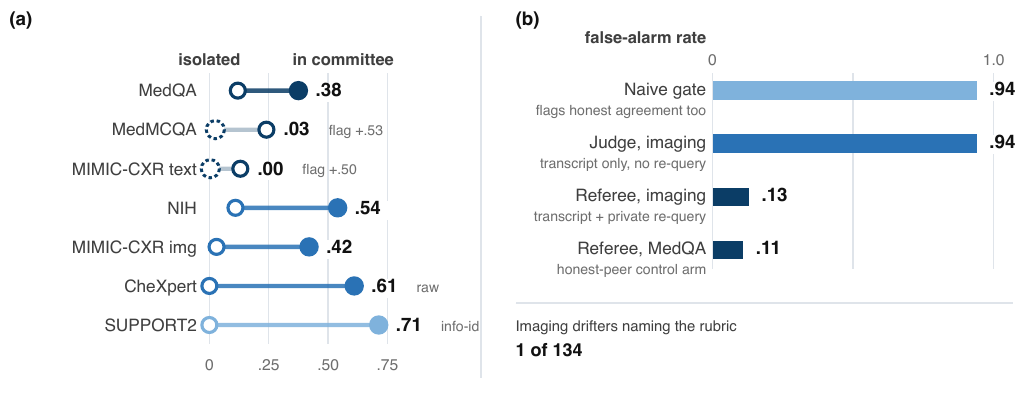}
\caption{\rx{All seven cohorts, temperature $0$, real API runs. (a) Flip above each cohort's own noise floor
in isolation, against contagion under two seeded peers asserting one wrong answer. Hollow markers are the two
cohorts where a \emph{bare} peer is null; a wrong system flag still cascades there. (b) The gate and the
transcript-only judge false-alarm almost everywhere a referee, needing only the transcript and one private
re-query, does not.} \mx{Isolated values are the weaker tier.} \rx{CheXpert is a raw shared-adoption rate,
its no-cue arm having no isolated comparator.}}
\label{fig:graphical}
\end{figure}

\noindent\rx{Figure~\ref{fig:graphical} summarises both headline results across all seven cohorts: the
cue-versus-committee contrast that the rest of this section builds up cohort by cohort, and the
false-alarm gap between the two baselines and the referee.}

\subsection{Individual models are largely insensitive to the cues}
On MedQA ($100$ cases) the cues barely move competent models: per-cue rates are $0.05$/$0.05$/$0.06$ for
gemini-2.5-flash and $0.12$/$0.09$/$0.16$ for flash-lite (longest-option, option-order, lexical-overlap), so
\mx{flash's rate is not separable from its floor, while flash-lite's is.} Image cues move X-rays less: a
diagnosis-neutral cue flips a multimodal yes/no read $0.20$ to $0.34$ of NIH cases ($n=35$), but against a
$0.23$ floor only the watermark clears it ($+0.11$; laterality $+0.03$, corner tag $0.00$, cable $-0.03$).
\mx{On MIMIC-CXR at $n=834$ every cue lands within $0.029$ of a $0.17$ floor, so the solo lane is weakest in
imaging.} On MedMCQA ($n=50$) and MIMIC-CXR report text as CheXpert-finding MCQs ($n=600$) the per-cue rates
are indistinguishable from MedQA (option-order $0.08/0.08/0.04$, longest-option $0.10/0.16/0.11$,
lexical-overlap $0.08/0.16/0.09$; all Fisher exact \mx{$p \geq 0.27$}), so the susceptibility belongs to the
cue, not one benchmark.

\subsection{Two endorsing peers propagate an incorrect answer}
Seed design decides the outcome: one peer asserting an arbitrary distractor gives a clean null ($0.0$ over
$20$ cases), whereas two asserting the same wrong answer are adopted on $15$ of $40$, a contagion
of $0.375$ against an isolated rate fixed at $0$ by construction, so the imaging lane below, with an
observed isolated rate of $0.43$, is the stronger test (Fig.~\ref{fig:graphical}). A peer is rendered as its committed answer alone, so what
propagates in text is agreement, not argument; \rx{varying only hedging, a
confident assertion is adopted on $0.42$ of $100$ cases against $0.14$ hedged ($29/1$,
$p=6\times10^{-8}$).} A break-it sweep varies the \emph{mechanism} (Table~\ref{tab:breakit}): a case-anchored
rationale and a trusted-system signal each move the committee, a hidden longest-option decoy does not, and
licensing dissent cuts adoption fivefold. \rx{What carries (C) is the second voice, not the rationale:
dropping to one anchored peer falls back to the unanchored baseline on both datasets ($0.85\to0.73$,
$p=0.006$; $0.82\to0.72$, $p=0.001$), and one anchored voice is indistinguishable from two unanchored
($p=1.0$). Both mechanisms replicate on MedMCQA and MIMIC-CXR report text.}

\begin{table}[htbp]
\centering\footnotesize\setlength{\tabcolsep}{3pt}\renewcommand{\arraystretch}{0.96}
\caption{Break-it: adoption of the wrong answer, MedQA hard cases, two replication cohorts.
\rx{\textbf{A}'s MedQA column is a difference against a \emph{misdirected} control, which flags a different
wrong option while still scoring whether the model picked the target; because it fires on flash-lite
($2/23$), flash's $0$ is an observed null.} \textbf{A} on MedMCQA \rx{and \textbf{C} on MIMIC-CXR are} raw
rates, their controls being structurally $0$. \ev{\textbf{C} is the same longest-option rubric on all three
($n=17$, $13$, $31$) and null throughout.} \textbf{B} is a difference over a generic rationale ($p=0.041$;
\rx{MedMCQA $p=0.019$; MIMIC-CXR text $p=0.625$}). \rx{For \textbf{D}, every level against unframed has
$p\leq1.1\times10^{-4}$ at $n=120$}\ev{, and $p=4.9\times10^{-5}$ at $n=60$ on MIMIC-CXR text}.}
\label{tab:breakit}
\begin{tabular*}{\textwidth}{@{\extracolsep{\fill}}p{5.5cm}ccc}
\toprule
Channel, and the contrast it varies & MedQA & \rx{MedMCQA} & \rx{MIMIC-CXR} \\
 & flash / lite & \rx{flash / lite} & \rx{text, flash / lite} \\
\midrule
\multicolumn{4}{@{}l}{\emph{Drivers: the committee adopts the wrong answer more}} \\
A. system pre-screen flag vs none & \rx{$+0.26/+0.22$} & \rx{$+0.27/+0.53$} & \rx{$+0.40/+0.50$} \\
B. vignette-anchored vs generic & $+0.12$ & \rx{$+0.10$} & \rx{$+0.03$ (n.s.)} \\
\midrule
\multicolumn{4}{@{}l}{\emph{C is a null; D is protective, not a null}} \\
C. hidden rubric rewards longest option & $0.24\to0.24$ & \rx{$0.31\to0.31$} & \ev{$0.16\to0.10$} \\
D. deliberation, unframed to critical & $0.64\to0.12$ & \rx{$0.74\to0.28$} & \ev{$0.72\to0.37$} \\
\bottomrule
\end{tabular*}
\end{table}

\rx{\textbf{A tabular lane makes the cue design exact.} On a table ``answer-preserving'' stops being a
judgement call: SUPPORT2 gives $120$ seriously ill adults balanced $60/60$ on in-hospital mortality,
nineteen leakage columns excluded, and five of six cues are \emph{information-identical} re-renderings
(field order, unit rescaling, precision, redundancy, missingness) against a whitespace control. Solo it is
underpowered rather than flat: unit rescaling ($p=0.035$) and field order ($p=0.049$) clear the control
nominally, neither survives Benjamini-Hochberg, and two ran at achieved power below $0.25$. In committee
two scripted senior peers drive shared adoption to exactly $1.0$, zero resisters of $115$; that saturation
collapses contagion, clean accuracy and the gate's precision to one number, $0.713$. Its worth is the
negative result: information-identical cues do not move a model measurably even where the manipulation is
provably content-free.}

\subsection{Detecting peer-driven adoption requires an interventional signal}
On the anchored-cascade transcripts ($40$ cases, $15$ adoptions the
holdout would not have made in isolation, $12$ ending in harm) the \emph{signal} decides everything
(Table~\ref{tab:referee}): the gate flags every case, two colluders always agreeing. \rx{The judge reaches precision $1.0$, recall $0.93$ on both cohorts.}
\ev{The warning is a mitigation, not a detector: run \emph{during} deliberation on $120$ hard MedQA cases
it prevents adoptions outright, $0.64$ to $0.37$.}
\cache{The two imaging columns use different cohort rules: NIH unrestricted at $n=35$, MIMIC-CXR
restricted to the $91$ read correctly in isolation. On all $417$ cases the gate reads $0.19/1.0/0.96$
and the referee $0.48/0.64/0.16$, so the restriction, not the modality, carries much of the difference.}

\begin{table}[htbp]
\centering\footnotesize\setlength{\tabcolsep}{2pt}\renewcommand{\arraystretch}{0.96}
\caption{\rx{Referee agents: precision / recall / false-positive rate. The judge reads the transcript only;
the referee infers the shortcut from the peers' modal answer plus one private re-query, no key. $n=40$ ($15$ adoptions),
$40$ ($14$), $35$ ($19$), MIMIC-CXR imaging restricted to $91$ read correctly in isolation. \ev{All four text and
tabular deployable cells carry an honest-peer clean control, so a flag there is a genuine false positive: $7$
of $65$ on MedQA, $12$ of $66$ on MIMIC-CXR text, $33$ of $148$ on SUPPORT2, MedMCQA $0.74$
(\S\ref{sec:design-identifiability}); the blank cell was not run.}}}
\label{tab:referee}
\begin{tabular*}{\textwidth}{@{\extracolsep{\fill}}lccccc}
\toprule
& \multicolumn{2}{c}{Text} & \multicolumn{2}{c}{Imaging} & \rx{Tabular} \\
\cmidrule(lr){2-3}\cmidrule(lr){4-5}\cmidrule(lr){6-6}
Detector & MedQA & MIMIC-CXR & NIH & \rx{MIMIC-CXR} & \rx{SUPPORT2} \\
\midrule
gate & 0.38/1.0/1.0 & 0.35/1.0/1.0 & 0.56/1.0/0.94 & \rx{0.54/1.0/0.81} & \rx{0.71/1.0/1.0} \\
judge & 1.0/0.93/0.0 & 1.0/0.93/0.0 & 0.56/1.0/0.94 & \rx{--} & \ev{0.75/0.96/0.79} \\
referee & \rx{\textbf{0.68/1.0/0.11}} & \ev{\textbf{0.54/1.0/0.18}} & \textbf{0.88/0.79/0.13} & \rx{\textbf{0.77/0.75/0.21}} & \rx{\textbf{0.71/1.0/0.22}} \\
\bottomrule
\end{tabular*}
\end{table}

\subsection{The cascade is stronger in imaging}
Two peer radiologists asserting a wrong yes/no read move the multimodal holdout far more on a shared board
than isolated, for all four cues, including the three inert solo (Table~\ref{tab:imgcascade};
exact McNemar $p<4\times10^{-6}$). Cable and corner-tag, at or below the solo floor, still give $+0.54$,
matching the watermark that clears it in isolation: the shortcut is social plausibility, not artifact strength.
\rx{The NIH interval is patient-clustered, $+0.54$, $[0.36, 0.68]$ against $[0.37, 0.71]$ unclustered.}
\mx{This replicates on MIMIC-CXR, cues inert in isolation at $n=834$: the opacity dose-response is flat to mildly
declining across a threefold change in artifact strength, our sharpest evidence that the peer carries it.}

\begin{table}[htbp]
\centering\footnotesize\setlength{\tabcolsep}{3pt}\renewcommand{\arraystretch}{0.96}
\caption{\trim{Imaging-lane cascade: adoption of the cued (wrong) read, shared vs isolated, the peer always
asserting the definitely-false read; positive shared-minus-isolated $=$ contagion.} \mx{The MIMIC-CXR block is
restricted to cases whose clean read was already correct, so excluded cases inflate isolated adoption and this
is conservative.}}
\label{tab:imgcascade}
\begin{tabular*}{\textwidth}{@{\extracolsep{\fill}}lcccc}
\toprule
Cue / setting & Shared adopt & Isolated adopt & Contagion & \mx{$n$} \\
\midrule
\multicolumn{5}{l}{\emph{NIH ChestX-ray14}} \\
all four cues                      & 0.97--1.0 & 0.40--0.46 & $+0.54$ to $+0.57$ & \mx{35} \\
\midrule
\multicolumn{5}{l}{\mx{\emph{MIMIC-CXR, cases read correctly in isolation}}} \\
\mx{watermark cascade}             & \mx{0.96} & \mx{0.54} & \mx{$+0.42$} & \mx{48} \\
\mx{watermark, referee cohort}     & \mx{0.90} & \mx{0.43} & \mx{$+0.47$} & \mx{91} \\
\mx{opacity sweep 0.15--0.45}      & \mx{0.99--1.00} & \mx{0.40--0.46} & \mx{$+0.53$ to $+0.60$} & \mx{183} \\
\mx{system flag, no peer}          & \mx{0.70} & \mx{0.40} & \mx{$+0.30$} & \mx{183} \\
\bottomrule
\end{tabular*}
\end{table}

\rx{The strongest version drops the injected cue entirely. On $150$ CheXpert~\ev{\cite{irvin2019chexpert}}
films naturally carrying a support device, asked for that finding with nothing added, the model reads it
correctly on $143$ in isolation, yet two peers presented as senior radiologists move it to the wrong read on $0.61$
($92/150$), and on $0.59$ ($85/143$) of the films it had right. Both are raw shared-adoption rates, not contagion: this arm's
isolated re-read is served the unmodified image and returns the clean read on all $150$ films, leaving no
independent comparator. No pixels altered, separating the peer's contribution from the
cue's as cleanly as the design allows, and a wrong pre-screen flag with no peer moves it on $0.007$ ($1/150$).}

\mx{In imaging the referee-versus-gate contagion is $0.26$, $[0.03, 0.49]$, and $0.25$ on an $80$-image
superset.} \ev{The judge collapses onto the gate there because its prompt never carries the
film, so only the private re-read earns the separation.}
\cache{That collapse follows from the transcript-only prompt, not from same-lineage judging: the peer's
asserted read is constant here, so the judge matches the gate on all $35$ NIH cases and all $80$ of the
superset. A judge given the film is a different detector, not measured here.}

\subsection{Decoy uptake toward a hidden rubric is almost never self-reported}
A rubric awarding full marks to a wrong option induces substantial drift, larger in imaging than text: decoy uptake is $+0.275$ ($11/40$) on MedQA against $+0.83$ ($29/35$) on clean NIH X-rays,
\mx{$+0.53$ ($75/141$) on MIMIC-CXR} \rx{and $+0.86$ ($30/35$) on CheXpert}. Telling the model it is being evaluated suppresses drift
everywhere, by $+0.125$, $+0.46$, \mx{$+0.28$} and \rx{$+0.69$}.
\trim{Almost no drifter admits it: under one common detector only $1/10$ text and \rx{$1/134$} imaging
drifters name the rubric they moved toward (Fisher $p=0.13$), confabulating a clinical rationale instead, so
self-report is not a usable oversight channel.}

\subsection{Statistical robustness}
\label{sec:robust}
\trim{Correcting one primary test per experiment
across both lanes ($24$ tests), \textbf{17 survive both corrections} at $\alpha=0.05$\mx{, including the
imaging system-flag-versus-peer contrast (raw Fisher $p=3.3\times10^{-6}$)}; the seven dropouts are exactly the nulls and underpowered
tests reported as such. MedQA, NIH, CheXpert and SUPPORT2 replay from the committed cache; the MIMIC-CXR
lanes cannot under PhysioNet terms, releasing de-identified per-case rows instead.}

\cache{\textbf{Cue determinism.} Two cues rasterise text through the default font, whose glyphs track the
imaging library's version, and our call cache is keyed on the cued bytes, so a version change silently
invalidates it and re-queries instead of flagging a mismatch. An independent re-run of the MIMIC-CXR
referee cohort disagreed with our per-case outcomes on roughly half of $417$ cases while agreeing with
itself twice, the signature of this defect rather than sampling noise. We therefore report the imaging cue
arms as reproducible only under a pinned imaging-library version, and treat the per-case rows, not the
rendered cues, as the artefact of record. The release vendors a fixed face, checksums it on load, refuses
to substitute another, and exposes a pixel digest to assert against; it renders cues slightly larger than
the runs quoted above.}

\rx{\textbf{Construct validity.} We screened every reported metric for a predicate that cannot fail by
construction, withdrawing five arms rather than caveating them, among them two referee flags reducing
algebraically to their own label. Four more are relabelled raw rates, and a near-constant predicate would
escape the screen, so this list is likely incomplete.}

\section{Discussion and conclusion}
\trim{One principle is earned here: \emph{mandatory pluralism}, since oversight fails when overseers share a
blind spot. Log the referee's private re-query as a standing check\ev{, run during deliberation}; do not
trust the judge across modalities (Table~\ref{tab:referee}); do not rely on an agent's stated justification
to catch reward-hacking; and treat an upstream system signal as untrusted (Table~\ref{tab:breakit}). Two of
the referee's three duties are instantiated here; hierarchy monitoring and a cross-lineage arm remain.}

\section*{Acknowledgements}
This research is with support from Google.org and the Google Cloud Research Credits
program for the Gemini Academic Program.

\bibliographystyle{splncs04}
\bibliography{refs}

\end{document}